\documentclass[runningheads, envcountsame, a4paper]{llncs}

\usepackage[ruled, vlined]{algorithm2e}

\usepackage{xcolor}
\usepackage[pdftex]{graphicx}
\usepackage{epstopdf}
\usepackage{subcaption}
\usepackage{amsmath}
\usepackage{amsfonts}
\usepackage{algpseudocode}
\usepackage{varwidth}
\usepackage{multirow}
\usepackage[hyphens]{url}
\usepackage{hyperref}
\begin{document}
\title{Anomaly Detection: How to Artificially Increase your F1-Score with a Biased Evaluation Protocol}
\titlerunning{Anomaly Detection: F1-score is biased}
% If the paper title is too long for the running head, you can set
% an abbreviated paper title here
%
\author{
Damien Fourure\thanks{alphabetical order}\orcidID{0000-0001-5085-0052}
\and
Muhammad Usama Javaid*\orcidID{0000-0001-9262-2250}%\inst{1}\orcidID{1111-2222-3333-4444} 
\and
Nicolas Posocco*\orcidID{0000-0002-1795-6039}%\inst{1}\orcidID{2222--3333-4444-5555} 
\and
Simon Tihon*\orcidID{0000-0002-3985-1967}%\inst{1}\orcidID{2222--3333-4444-5555}
}

\authorrunning{D. Fourure et al.}
\institute{EURA NOVA, Mont-St-Guibert, Belgium\\
\email{$\{$firstname$\}$.$\{$lastname$\}$@euranova.eu}}

%
% \authorrunning{F. Author et al.}
% First names are abbreviated in the running head.
% If there are more than two authors, 'et al.' is used.
%
% \institute{Princeton University, Princeton NJ 08544, USA \and
% Springer Heidelberg, Tiergartenstr. 17, 69121 Heidelberg, Germany
% \email{lncs@springer.com}\\
% \url{http://www.springer.com/gp/computer-science/lncs} \and
% ABC Institute, Rupert-Karls-University Heidelberg, Heidelberg, Germany\\
% \email{\{abc,lncs\}@uni-heidelberg.de}}
%
\maketitle  % typeset the header of the contribution
\setcounter{footnote}{0}

\begin{abstract}
Anomaly detection is a widely explored domain in machine learning. Many models are proposed in the literature, and compared through different metrics measured on various datasets.
The most popular metrics used to compare performances are F1-score, AUC and AVPR.
In this paper, we show that F1-score and AVPR are highly sensitive to the contamination rate.
One consequence is that it is possible to artificially increase their values by modifying the train-test split procedure.
This leads to misleading comparisons between algorithms in the literature, especially when the evaluation protocol is not well detailed.
Moreover, we show that the F1-score and the AVPR cannot be used to compare performances on different datasets as they do not reflect the intrinsic difficulty of modeling such data.
Based on these observations, we claim that F1-score and AVPR should not be used as metrics for anomaly detection. We recommend a generic evaluation procedure for unsupervised anomaly detection, including the use of other metrics such as the AUC, which are more robust to arbitrary choices in the evaluation protocol.
\end{abstract}

\section{Introduction}

Anomaly detection has been widely studied in the past few years, mostly for its immediate usability in real-world applications. Though there are multiple definitions of anomalies in the literature, most definitions agree on the fact that anomalies are data points which do not come from the main distribution. In the setting of unsupervised anomaly detection, the goal is to create a model which can distinguish anomalous samples from normal ones without being given such label at train time. In order to do so, most approaches follow a one-class classification framework, which models the normal data from the train set, and predicts as anomalous any point which does not fit this distribution of normal samples. Such prediction needs some prior knowledge provided through a contamination rate on the test set, which is the ratio of anomalous data within. This ratio is used to build the model's decision rule. 

In this setting, a lot of the literature uses the F1-score or the average precision (AVPR) to evaluate and compare models. In this paper we show that the evaluation protocol (train-test split and contamination rate estimation) has a direct influence on the contamination rate of the test set and the decision threshold, which in turn has a direct influence on these metrics. We highlight a comparability issue between results in different papers based on such evidence, and suggest an unbiased protocol to evaluate and compare unsupervised anomaly detection algorithms.

After an extensive study of the unsupervised anomaly detection field and of previous analyses of the evaluation methods (Section~\ref{sec:related_works}), we study the impact of the evaluation procedure on commonly used metrics (Section~\ref{sec:impact_setup_f1_score}). Identified issues include a possibility to artificially increase the obtained scores and a non-comparability of the results over different datasets. Taking these into account, we suggest the use of a protocol leading to a better comparability in Section~\ref{sec:proposed_alternative}.

\section{Related Work}
\label{sec:related_works}

Anomaly detection has been heavily dominated by unsupervised classification settings. One very popular approach in unsupervised anomaly detection is one-class classification, which refers to the setting where at train time, the model is given only normal samples to learn what the normal distribution is. The goal is to learn a scoring function to assign each data point an abnormality score. A threshold is then calculated from either a known or estimated contamination rate to turn scores into labels, samples with higher scores being considered as anomalies. In the literature different scoring functions have been used:

\paragraph{Proximity-based methods} use heuristics based on distances between samples in some relevant space. These algorithms estimate the local density of data points through distances, and point out the most isolated ones. Legacy approaches include a simple distance to the Kth neighbour \cite{KNN}, Angle-Based Outlier Detection (ABOD) \cite{ABOD}, which uses the  variance over the angles between the different vectors to  all pairs of points weighted by the distances between them, Local Outlier Factor (LOF) \cite{LOF}, which measures the local deviation of a given data point with respect to its neighbours, Connectivity-based Outlier Factor (COF) \cite{COF}, which uses a ratio of averages of chaining distances with neighbours and Clustering-Based Local Outlier Factor (CBLOF) \cite{CBLOF}, which clusters the data and scores samples based on the size of the cluster they belong to and the distance to the closest big cluster. More recent approaches include DROCC \cite{DROCC}, which makes the assumption that normal points lie on a well-sampled, locally linear low dimensional manifold and abnormal points lie at least at a certain distance from this manifold.

\paragraph{Reconstruction-based} approaches use notions of reconstruction error to determine which data points are anomalous, the reconstruction of the densest parts of the distribution being easier to learn in general. In \cite{PCA} for instance, the projection of each point on the main PCA axes is used to detect anomalies. As for \cite{yang2020memory}, a GAN with a memory matrix is presented, each row containing a memorised latent vector with the objective to enclose all the normal data, in latent space, in between memorised vectors. The optimisation introduces a reconstruction error.

\paragraph{Representation-based} approaches attempt to project the data in a space in which it is easy to identify outliers. Following this idea, One-Class SVM (OC-SVM) \cite{OCSVM} uses a hypersphere to encompass all of the instances in the projection space. \cite{DBLP:journals/corr/abs-1904-00152} proposed a neural network with robust subspace recovery layer. IDAGMM \cite{IDAGMM} presents an iterative algorithm based on an autoencoder and clustering, with the hypothesis that normal data points form a cluster with low variance. OneFlow \cite{maziarka2020flowbased}, is a normalising-flow based method which aims at learning a minimum enclosing ball containing most of the data in the latent space, the optimisation ensuring that denser regions are projected close to the origin.

\paragraph{Adversarial scoring} use the output of a discriminator as a proxy for abnormality, since it is precisely the goal of a discriminator to distinguish normal samples from other inputs. Driven by the motivation, an ensemble gan method is proposed in \cite{han2020gan}. GANomaly \cite{Ganomaly} presents a conditional generative adversarial network with a encoder-decoder-encoder network to train better on normal images at training, and \cite{yang2020regularized} presents Adversarially Learned Interface method with cycle consistency to ensure good reconstruction of normal data in one-class setting. \cite{9157105} presents a gan network with autoencoder as generator for anomaly detection on images datasets.

\paragraph{Feature-level} approaches try to detect anomalies at feature-level, and aggregate such information on each sample to produce an abnormality score at sample level. HBOS \cite{HBOF} assumes feature independence and calculates the degree of abnormality by building histograms. RVAE \cite{rvae} uses a variational autencoder to introduce cell abnormality, which is converted into sample anomaly detection.\\

All of these categories are of course non-exclusive, and some approaches, as the very popular Isolation Forest \cite{IF}, which uses the mean depth at which each sample is isolated in a forest of randomly built trees, do not fall in any of these. On the opposite, some recent methods combine multiple of such proxies for abnormality to reach better performances, each one using different hypotheses to model anomalies. For example \cite{zong2018deep} presents an end-to-end anomaly detection architecture. The model uses an autoencoder to perform dimensionality reduction to one or two dimensions and calculates several similarity errors, feeding then both latent representation and reconstruction errors to the gaussian mixture model. AnoGAN \cite{AnoGAN}, which uses both a reconstruction error and a discriminator score to detect anomalies, also falls in this category. \\

Even if the original one-class setting requires data to be all normal at train time (which makes one-class approaches not strictly unsupervised) some approaches do not require clean data at train time, since they use what they learn about normal data to reduce as much as possible the impact of anomalies \cite{IDAGMM,maziarka2020flowbased}. \\

For all these settings, the main evaluation metrics used in the literature are the  F1-score, the AUC (area under ROC curve) and the AVPR (average precision). The link between sensitivity, specificity and F1-score has been studied in \cite{lipton2014}, providing thresholding-related insights. In this work, we highlight the heterogeneity of current evaluation procedures in unsupervised anomaly detection performed in a one-class framework, would it be in terms of metrics or contamination-rate determination. For instance, many papers do not provide complete information about how the train-test splits are made \cite{rvae,maziarka2020flowbased}. For the same datasets, some papers re-inject the train anomalies in the test set \cite{mtq,zong2018deep,han2020gan}
\footnote{\cite{zong2018deep} do not publish their code but an unofficial implementation widely used (264 stars and 76 forks at the time of writing) is at available \url{https://github.com/danieltan07/dagmm}} and some others do not \cite{zong2018deep,DSEBM}. In some cases, it is not clear which contamination rate was used to compute the threshold \cite{9157105,maziarka2020flowbased,zong2018deep,han2020gan,yang2020memory}, and some approaches prefer evaluating their model with multiple thresholds \cite{DBLP:journals/corr/abs-1904-00152}. Different metrics are used to evaluate performences - F1-score \cite{9157105,yang2020regularized,DROCC,IDAGMM,han2020gan,zong2018deep}, precision \cite{yang2020regularized,han2020gan,zong2018deep}, recall \cite{yang2020regularized,han2020gan,zong2018deep}, sensitivity \cite{AnoGAN}, specificity \cite{AnoGAN}, AUC \cite{9157105,Ganomaly,yang2020regularized,yang2020memory,DROCC,AnoGAN,NEURIPS2019_6c4bb406,IDAGMM,Ergen_2020,DBLP:journals/corr/abs-1904-00152,han2020gan,rvae}, AVPR \cite{DBLP:journals/corr/abs-1904-00152,IDAGMM,rvae,NEURIPS2019_6c4bb406}. Finally many papers report directly the results from other papers and do not test the associated algorithms in their particular evaluation setting.

We show that all above-mentioned setup details have a direct impact on the F1-score and the AVPR. Since such heterogeneity leads to reproducibility and comparability issues, we suggest the use of an evaluation protocol with a robust metric which allows comparison.

\section{Issues when Using F1-Score and AVPR Metrics}
\label{sec:impact_setup_f1_score}

In this section, we analyse the sensitivity of the F1-score and AVPR metrics with respect to the contamination rate of the test set. First, we define the problem and different metrics and explain the impact of the estimation of the contamination rate. Then, we analyse the evolution of the metrics according to the true contamination rate of the test set. After having explained different evaluation protocols used in the literature, we show how they can be used to produce artificially good results using the F1-score and AVPR metrics. Finally, we show that these two metrics are also unsuitable for estimating the difficulty of datasets.

\subsection{Formalism and Problem Statement} \label{subsec:formalism}
Consider a dataset $\mathbf{D} = \{(\mathbf{x}_1, y_1),\ \ldots, (\mathbf{x}_N, y_N)\} \subset \mathbb{R}^d \times \{0, 1\}$, with $\mathbf{x}_i$ the $d$-dimensional samples and $y_i$ the corresponding labels. We assume both classes are composed of i.i.d. samples. We also assume the normal class labeled $0$ outnumbers the anomaly class labeled $1$. Therefore, we choose the anomaly class as positive class and use $^+$ to refer to it, while using $^-$ to refer to the normal class. This dataset is split into a train set $\mathbf{D}^{train} \subset \mathbf{D}$ and a test set $\mathbf{D}^{test} = \mathbf{D} \setminus \mathbf{D}^{train}$. Different procedures are used in the anomaly-detection community to perform this split, as detailed in Section~\ref{subsec:train-test-split}. We denote $N_t^+$ (resp. $N_t^-$) the number of anomalous (resp. normal) samples in the test set.

We consider one-class classifiers, which are models learning an anomaly-score function $f$ based only on clean samples $\mathbf{X}^{clean} = \{\mathbf{x}\ \forall (\mathbf{x}, y) \in \mathbf{D}^{train} \mid y = 0\}$. The anomaly-score function returns, for a given sample $\mathbf{x}$, an anomaly score $\hat{s} = f(\mathbf{x}) \in \mathbb{R}$ such that the higher the score, the more likely it is that $\mathbf{x}$ is an anomaly. We define $P^+(\hat{s})$ (resp. $P^-(\hat{s})$) the probability that an anomaly (resp. a clean sample) obtains an anomaly-score $\hat{s}$ with the trained model.

To get a binary prediction $\hat{y}$ for a sample $\mathbf{x}$ with anomaly score $\hat{s}$, we need to apply a threshold $t$ to the anomaly score such that $\hat{y} = 1 \text{ if } \hat{s} \geq t \text{ else } \hat{y} = 0$. Different ways to compute this threshold are used in the literature. A common approach is to choose it according to an estimation $\hat{\alpha}$ of the contamination rate $\alpha$. The contamination rate is the proportion of anomalous samples in the dataset. It can be taken as domain knowledge, estimated on the train set or, for evaluation purposes only, on the test set directly. 

\subsection{Definition of the Metrics}

\begin{figure}[t]
\centering
\begin{subfigure}{0.33\textwidth}
    \begin{tabular}{c|c|c}
        & Actual & Actual \\
        & Anomaly & Normal \\
        \hline
        Predicted & \multirow{2}{*}{$tp$} & \multirow{2}{*}{$fp$}\\
        Anomaly &  &  \\
        \hline
        Predicted & \multirow{2}{*}{$fn$} & \multirow{2}{*}{$tn$}\\
        Normal &  &  \\
    \end{tabular}
    \caption{Confusion Matrix}
    \label{subfig:conf_mat}
\end{subfigure}
\begin{subfigure}{0.66\textwidth}
    \begin{equation}
        \textit{precision} = \frac{tp}{tp + fp}
    \end{equation}
    \begin{equation}
        \textit{recall} = \frac{tp}{tp + fn}
    \end{equation}
    \begin{equation}
        \textit{F1-score} = \frac{2}{\textit{precision}^{-1} + \textit{recall}^{-1}}
    \end{equation}
    \caption{Metrics based on binary predictions}
    \label{subfig:F1_form}
\end{subfigure}
\caption{Metrics definitions.}
\end{figure}

\begin{figure}[t]
    \centering
    % ---------- Arrhythmia -----------------
    \begin{subfigure}{0.33\textwidth}
        \centering
        \includegraphics[width=\textwidth]{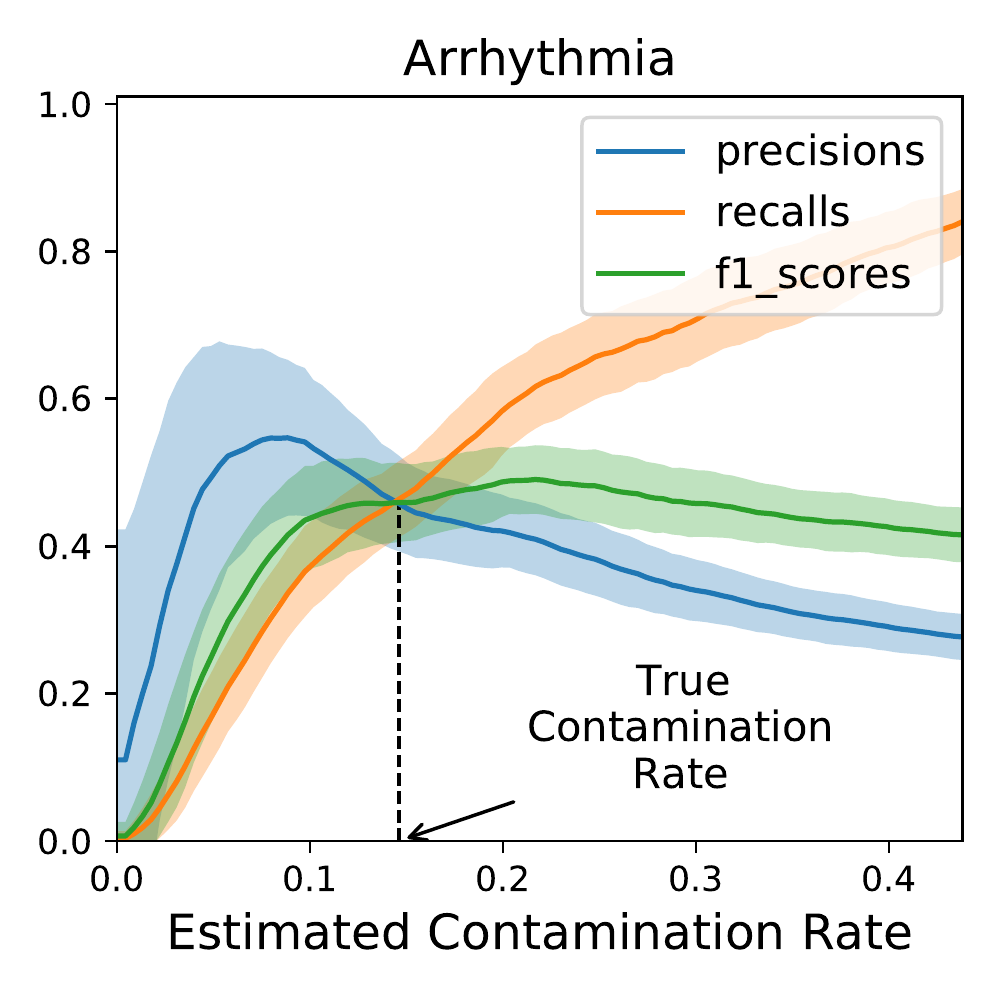}
    \end{subfigure}%
    % ---------- Thyroid -----------------
    \begin{subfigure}{0.33\textwidth}
        \centering
        \includegraphics[width=\textwidth]{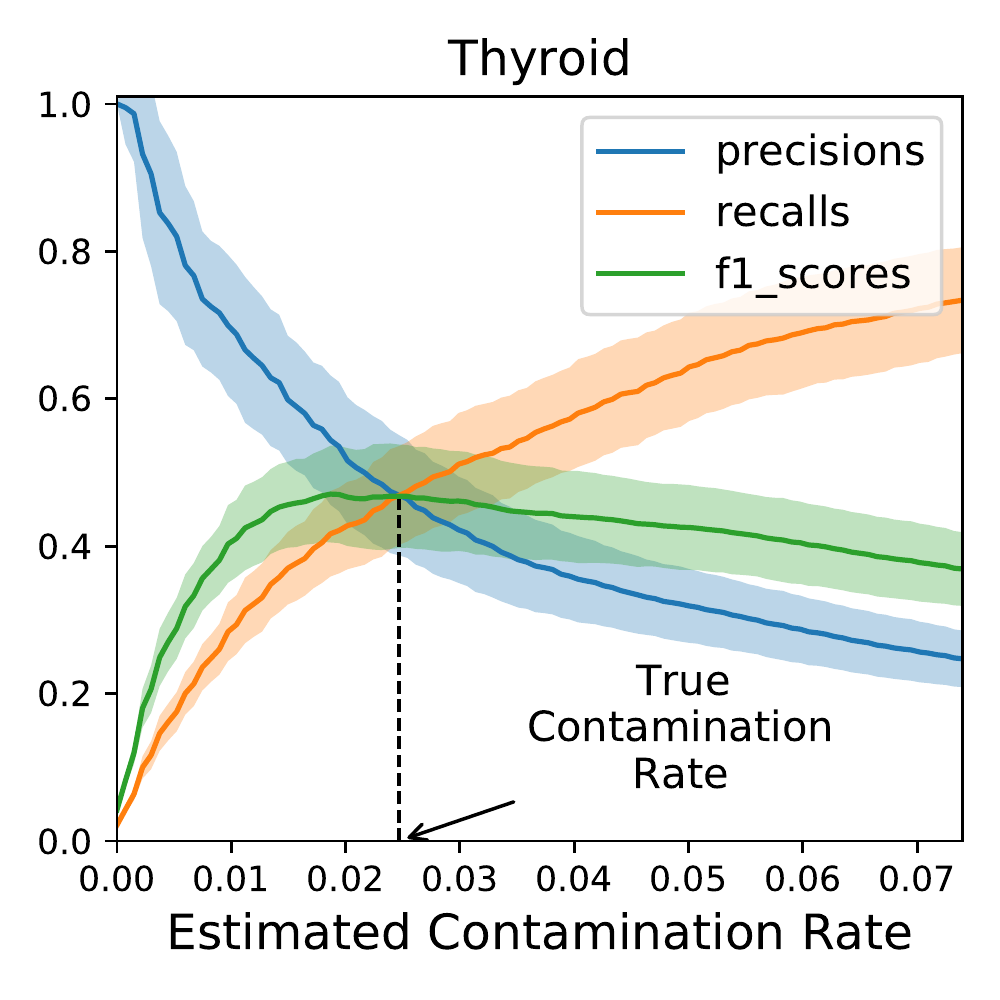}
    \end{subfigure}%
    % ---------- Kdd cup -----------------
    \begin{subfigure}{0.33\textwidth}
        \centering
        \includegraphics[width=\textwidth]{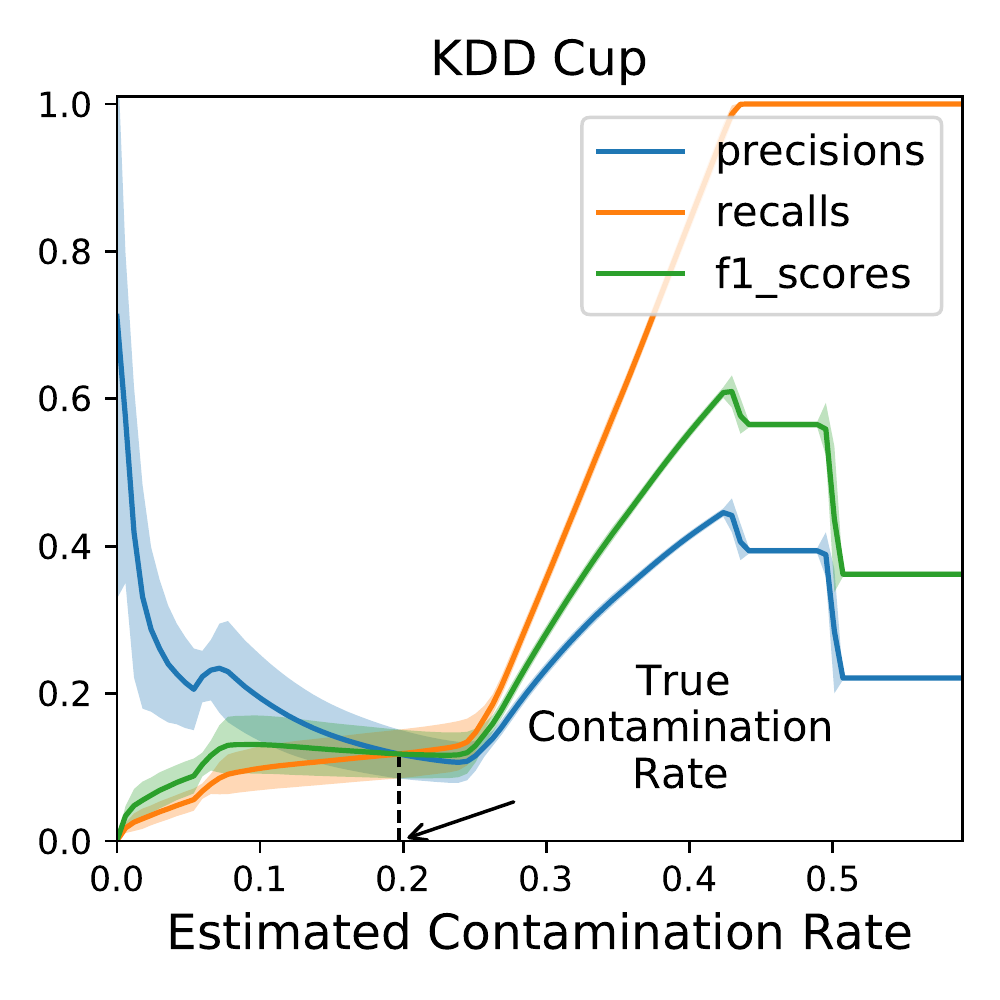}
    \end{subfigure}%
    \caption{Evolution of the \textit{Precisions}, \textit{Recalls} and \textit{F1-scores} according to the estimated contamination rate on three different datasets. The curves are obtained using the Algorithm~\ref{algo:theoretical} introduced in Section~\ref{subsec:train-test-split}.} 
    \label{fig:precision_recall_f1}
\end{figure}

Using the final prediction and the ground truth labels, we can count the \textit{true positives} $tp$, \textit{true negatives} $tn$, \textit{false positives} $fp$ and \textit{false negatives} $fn$, as shown in Figure~\ref{subfig:conf_mat}. The \textit{precision}, \textit{recall} and \textit{F1-score} are computed using these quantities as shown in the equations of Figure~\ref{subfig:F1_form}. An example of these metrics applied with a varying contamination rate estimation $\hat{\alpha}$, inducing a varying threshold, is shown in Figure~\ref{fig:precision_recall_f1}. It is interesting to note that, if the estimated contamination rate $\hat{\alpha}$ is equal to the true contamination rate $\alpha$, we have \textit{precision} = \textit{recall} = \textit{F1-score}. This can be easily explained: if the estimated contamination rate is the true contamination rate, the threshold is computed such that the number of samples predicted as anomalous is equal to the number of true anomalies in the set. Thus, if a normal sample is wrongly predicted as anomalous (i.e. is a false positive), it necessarily means that an anomalous sample has been predicted as normal (i.e. is a false negative). That is, $fp = fn$. Given the formulas of precision and recall (see equations of Figure~\ref{subfig:F1_form}) we have $precision = recall$. As the F1-score is the harmonic mean of precision and recall, we have $precision = recall = \textit{F1-score}$. Inversely, if this equality can be observed in reported results, it is safe to assume the estimation of the contamination rate is equal to the true contamination rate.

\begin{figure}[b]
    \centering
    \begin{subfigure}{0.5\textwidth}
        \centering
        \includegraphics[width=0.66\textwidth]{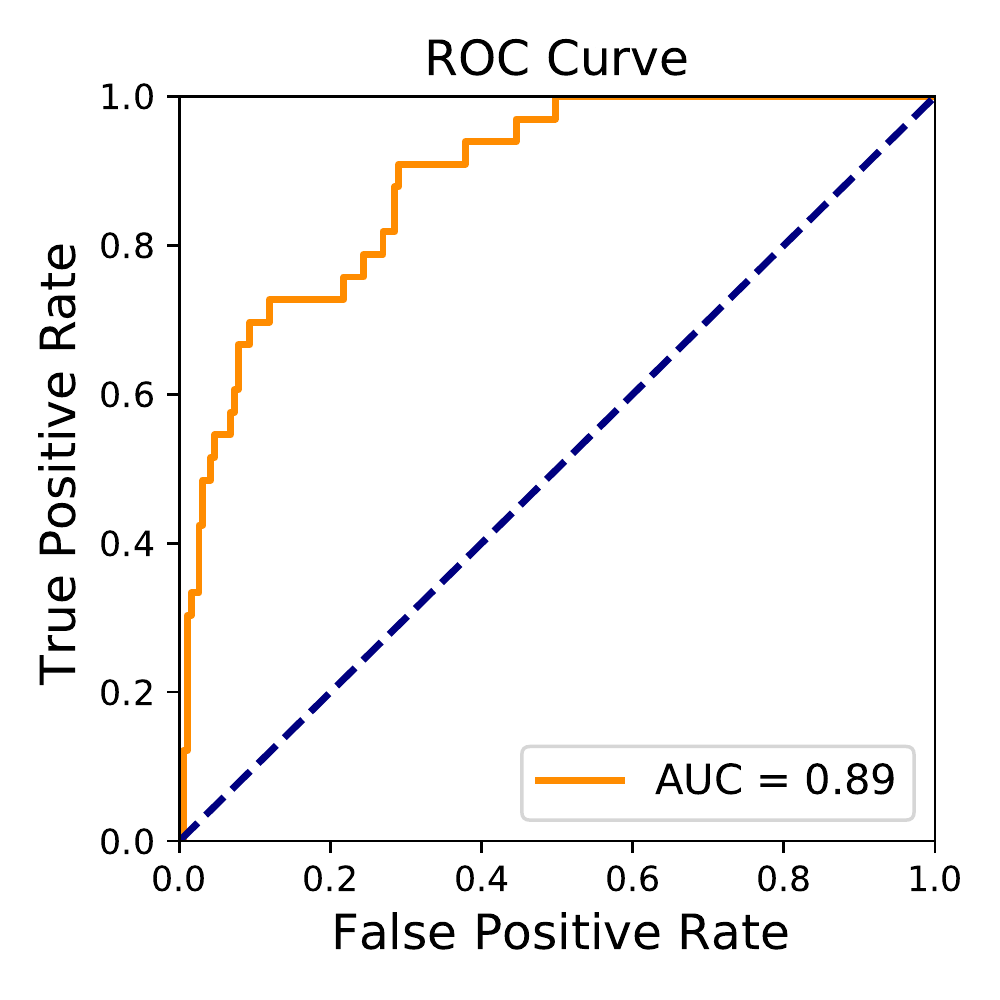}
    \end{subfigure}%
    \begin{subfigure}{0.5\textwidth}
        \centering
        \includegraphics[width=0.66\textwidth]{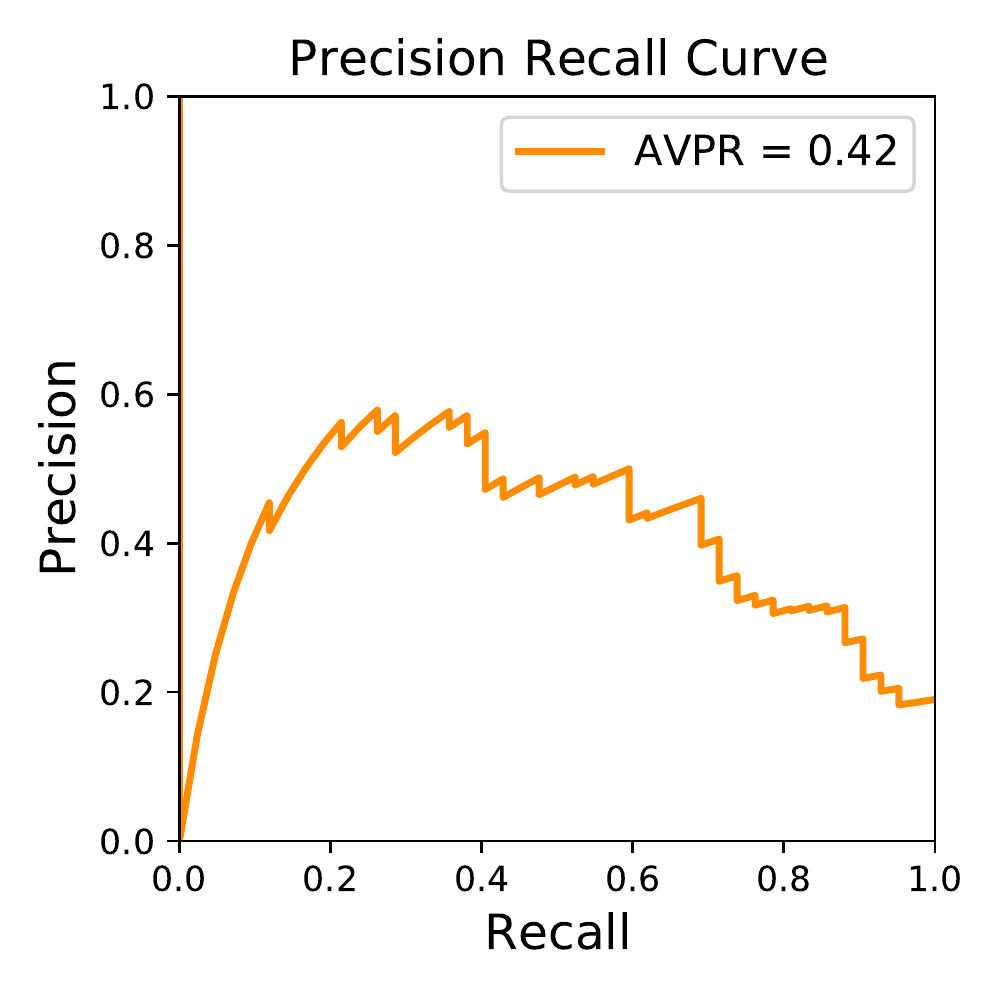}
    \end{subfigure}
    \caption{Example of ROC Curve and Precision Recall Curve obtained on the Arrhythmia dataset. The scores are obtained using the Algorithm~\ref{algo:theoretical} introduced in Section~\ref{subsec:train-test-split}.} 
    \label{fig:roc_pr_curve}
\end{figure}

We also include the AUC and AVPR in our analysis. These metrics are obtained by analysing the results with different thresholds. The AUC is defined through the receiver-operator characteristic (ROC) curve, a curve of the true positive rate over the false positive rate for various thresholds. Therefore, we redefine $tp$, $fp$, $fn$ and $tn$ as functions depending on the threshold. The area under the ROC curve \textit{AUC}, sometimes written \textit{AUROC}, is the total area under this curve, that is:
\begin{equation} \label{eq:auc}
    AUC = \int_{t=-\infty}^{\infty} \frac{tp(t)}{tp(t) + fn(t)} \frac{d}{dt}\left(\frac{fp}{fp+tn}\right)\Bigr\rvert_{t} dt.
\end{equation}

Similarly, the AVPR is defined through the precision-recall (PR) curve, a curve of precision over recall for different thresholds. The area under this curve is referred to as the average precision (AVPR) metric, as it can be seen as a weighted average of the precision for different recall. We have 
\begin{equation} \label{eq:avpr}
    AVPR = \int_{t=-\infty}^{\infty} precision(t) \frac{d}{dt}(recall)\Bigr\rvert_{t}dt.
\end{equation}
An example of a ROC curve and a precision recall curve is given in Figure~\ref{fig:roc_pr_curve}. 

We show in this paper that the F1-score and AVPR metrics are highly sensitive to the true contamination rate of the test set. We show this sensitivity has a negative impact on the comparison of different classifiers or datasets, especially when using different protocols.

\subsection{Evaluation Protocols: Theory vs Practice} \label{subsec:train-test-split}
Machine learning theory tells us that the evaluation of an algorithm should be done on a test set completely separated from the train set. Algorithm~\ref{algo:theoretical} presents the unbiased procedure to train and evaluate an anomaly detection model. A dataset (containing both normal and anomalous samples) is split into a train set and a test set. The anomalous samples from the train set are removed to get a clean set that is used to train a model. The train set is also used to compute the contamination rate and fix the threshold, for example using a threshold such that the train set has as many anomalies as predicted anomalies, i.e. $fp = fn$. This threshold is finally used on the predictions made on the new (unseen) samples composing the test set to measure the F1-score. The AUC and AVPR are computed using the predicted scores directly. Even though this procedure is theoretically the correct way to evaluate a model, it has a significant drawback in practice. The anomalous samples in the train set are used only to compute the threshold for the F1-score and are then thrown away. Because there are, by definition, few anomalies in a dataset, one could be tempted to use these samples in the test set. Indeed, as visible in Figure~\ref{fig:accuracy_per_test_size}, the more anomalous samples we can use to evaluate a model, the more precise the evaluation.

\begin{algorithm}[ht!]
    \SetAlgoLined
    \textbf{Input:}\par
    
    \hskip\algorithmicindent $\mathbf{D} \subset \mathbb{R}^d \times \{0, 1\}$ a set of $N$ $d$-dimensional input samples and their corresponding labels ($1=$ anomaly, $0=$ normal)\par
    
    \hskip\algorithmicindent $\beta$ the amount of data used for the test set\par
    
    \hskip\algorithmicindent $f$ a trainable anomaly-score function\par
    
    \textbf{Output:}\par
    \hskip\algorithmicindent F1-score, AUC and AVPR
    
    \textbf{Procedure:}\par
    \hskip\algorithmicindent $\mathbf{D}^{train}, \mathbf{D}^{test} = \text{split\_train\_test} (\mathbf{D}$, $\beta$)\par
    
    \hskip\algorithmicindent $\mathbf{X}^{clean} = \{\mathbf{x}\ \forall (\mathbf{x}, y) \in \mathbf{D}^{train} \mid y = 0\}$ \par
    
    \hskip\algorithmicindent Normalise the data based on $\mathbf{X}^{clean}$ if necessary
    
    \hskip\algorithmicindent Train $f$ using $\mathbf{X}^{clean}$
    
    \hskip\algorithmicindent $\mathbf{\hat{s}}^{train} = \{(f(\mathbf{x}), y)\ \forall (\mathbf{x}, y) \in \mathbf{D}^{train}\}$ \par
    
    \hskip\algorithmicindent Compute estimated contamination rate $\hat{\alpha} = \frac{|\{(\mathbf{x}, y)\ \forall (\mathbf{x}, y) \in \mathbf{D}^{train} \mid y = 1\}|}{|\mathbf{D}^{train}|}$\par
    
    \hskip\algorithmicindent Compute threshold $t$ such that $ \frac{|\{(\hat{s}, y)\ \forall (\hat{s}, y) \in \mathbf{\hat{s}}^{train} \mid \hat{s} \geq t\}|}{|\mathbf{\hat{s}}^{train}|} = \hat{\alpha}$ \par
    
    \hskip\algorithmicindent $\mathbf{\hat{s}}^{test} = \{(f(\mathbf{x}), y)\ \forall (\mathbf{x}, y) \in \mathbf{D}^{test}\}$ \par
    
    \hskip\algorithmicindent $\mathbf{\hat{y}}^{test} = \{(\hat{y}, y)\ \forall (\hat{s}, y) \in \mathbf{\hat{s}}^{test}\ \forall \hat{y} \in \{0, 1\} \mid  \hat{y} = 1 \text{ if } \hat{s} \geq t\text{ else }  \hat{y}=0\}$ \par
    
    \hskip\algorithmicindent Compute F1-score using $\mathbf{\hat{y}}^{test}$
    
    \hskip\algorithmicindent Compute AUC and AVPR using $\mathbf{\hat{s}}^{test}$
    
    \caption{Theoretically unbiased evaluation protocol}
    \label{algo:theoretical}
\end{algorithm}

To make full use of the anomalous samples, the procedure described in Algorithm~\ref{algo:recycling} recycles the anomalous samples contained in the train set. The threshold is then computed on the test set as there are no anomalies left in the train set to estimate it. This leads to a situation where $precision = recall = \textit{F1-score}$ as described in Section~\ref{subsec:formalism}. 
This recycling procedure makes sense in the context of anomaly detection as it obtains more precise results, and can be found in the literature \cite{mtq,zong2018deep}.

\begin{algorithm}[ht!]
    \SetAlgoLined
    
    \textbf{Input:} \\
    \hskip\algorithmicindent $\mathbf{D} \subset \mathbb{R}^d \times \{0, 1\}$ a set of $N$ $d$-dimensional input samples and their corresponding labels ($1=$ anomaly, $0=$ normal)\par
    
    \hskip\algorithmicindent $\beta$ the amount of data used for the test set\par
    
    \hskip\algorithmicindent $f$ a trainable anomaly-score function\par

    \textbf{Output:}\par
    \hskip\algorithmicindent F1-score, AUC and AVPR
    
    \textbf{Procedure:}\par
    \hskip\algorithmicindent $\mathbf{D}^{train}, \mathbf{D}^{test} = \text{split\_train\_test} (\mathbf{D}$, $\beta$)\par
    
    \hskip\algorithmicindent $\mathbf{X}^{clean} = \{\mathbf{x}\ \forall (\mathbf{x}, y) \in \mathbf{D}^{train} \mid y = 0\}$ \par
    
    \hskip\algorithmicindent Add $\{(\mathbf{x}, y)\ \forall (\mathbf{x}, y) \in \mathbf{D}^{train} \mid y = 1\}$ to $\mathbf{D}^{test}$ \par
    
    \hskip\algorithmicindent Normalise the data based on $\mathbf{X}^{clean}$ if necessary
    
    \hskip\algorithmicindent Train $f$ using $\mathbf{X}^{clean}$
    
    \hskip\algorithmicindent $\mathbf{\hat{s}}^{test} = \{(f(\mathbf{x}), y)\ \forall (\mathbf{x}, y) \in \mathbf{D}^{test}\}$ \par
    
    \hskip\algorithmicindent Compute contamination rate $\alpha = \frac{|\{(\mathbf{x}, y)\ \forall (\mathbf{x}, y) \in \mathbf{D}^{test} \mid y = 1\}|}{|\mathbf{D}^{test}|}$\par

    \hskip\algorithmicindent Compute threshold $t$ such that $ \frac{|\{(\hat{s}, y)\ \forall (\hat{s}, y) \in \mathbf{\hat{s}}^{test} \mid \hat{s} \geq t\}|}{|\mathbf{\hat{s}}^{test}|} = \alpha$ \par

    \hskip\algorithmicindent $\mathbf{\hat{y}}^{test} = \{(\hat{y}, y)\ \forall (\hat{s}, y) \in \mathbf{\hat{s}}^{test}\ \forall \hat{y} \in \{0, 1\} \mid  \hat{y} = 1 \text{ if } \hat{s} \geq t\text{ else }  \hat{y}=0\}$ \par
    
    \hskip\algorithmicindent Compute F1-score using $\mathbf{\hat{y}}^{test}$
    
    \hskip\algorithmicindent Compute AUC and AVPR using $\mathbf{\hat{s}}^{test}$
    
    \caption{\textit{Recycling} evaluation protocol for anomaly detection}
    \label{algo:recycling}
\end{algorithm}

Algorithms~\ref{algo:theoretical} and \ref{algo:recycling} take as input any dataset and any trainable anomaly-score function. For the dataset, if not specified otherwise, we use the Arrhythmia and Thyroid datasets from the {\color{blue}\href{http://odds.cs.stonybrook.edu/about-odds/}{ODDS repository}}~\cite{ODDS} and the Kddcup dataset from the {\color{blue}\href{http://archive.ics.uci.edu/ml/index.php}{UCI repository}}~\cite{uci_kddcup}. These datasets are often used in the anomaly-detection literature, and are therefore all indicated for our analysis. They have respectively $452$, $3772$ and $494020$ samples, with a contamination rate of respectively $14.6\%$, $2.5\%$ and $19.7\%$. For Kddcup, as done in the literature, the samples labeled as \textit{"normal"} are considered as anomalous and, for computational reasons, only $10\%$ are used.
For the trainable anomaly-score function, if not specified otherwise, we use OC-SVM~\cite{OCSVM} with its default hyper-parameters, as implemented in \texttt{sklearn}~\cite{scikit-learn}.
We choose this model as it has proven its worth and is often used as a baseline in the literature. 
We run all our experiments $100$ times to report meaningful means and standard deviations.
The code to reproduce all our figures and results is available at \url{https://github.com/euranova/F1-Score-is-Biased}.%\footnote{The code will be publicly available on github, but, to ensure the double blind process, the link is hidden and the code is provided to reviewers in the supplementary material.}.

\subsection{Metrics Sensitivity to the Contamination Rate of the Test Set} \label{subsec:sensitivity}

\begin{figure}[t]
    \centering
    \begin{subfigure}{0.32\textwidth}
        \centering
        \includegraphics[width=\textwidth]{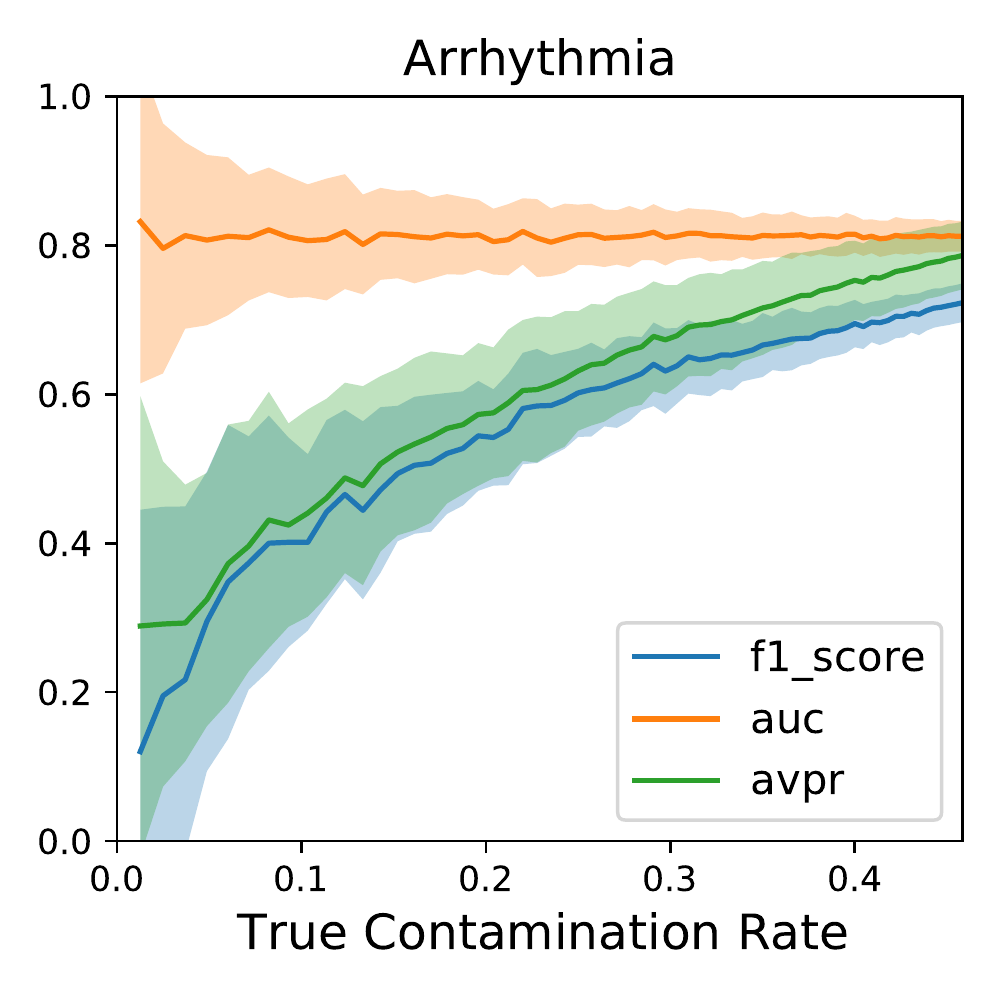}
    \end{subfigure}
    \begin{subfigure}{0.32\textwidth}
        \centering
        \includegraphics[width=\textwidth]{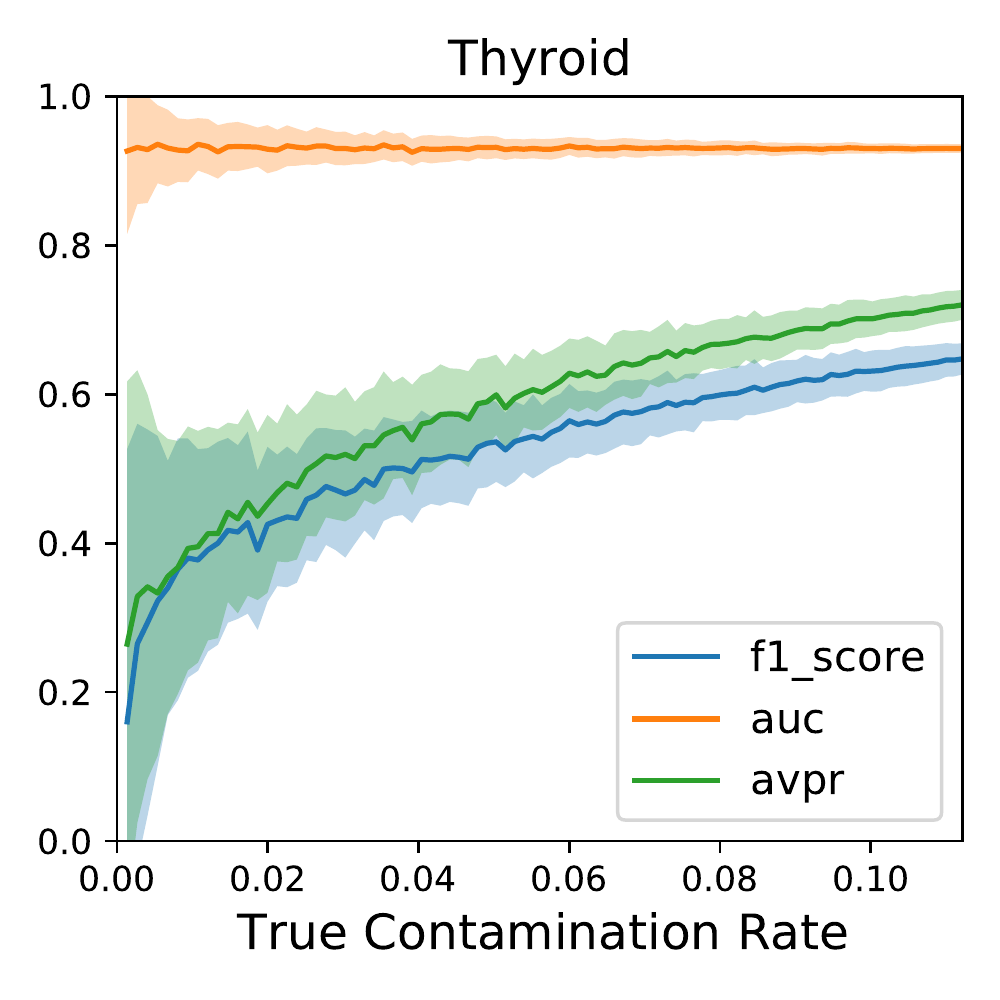}
    \end{subfigure}
    \begin{subfigure}{0.32\textwidth}
        \centering
        \includegraphics[width=\textwidth]{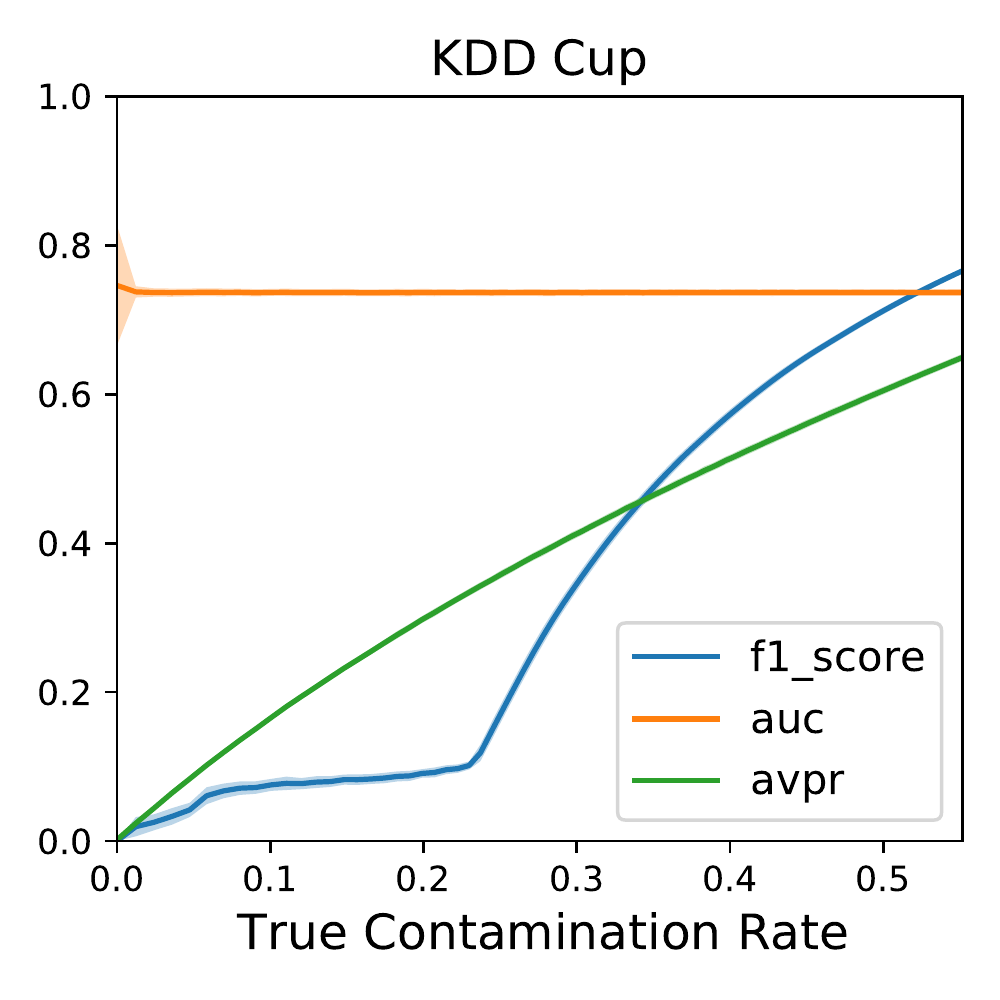}
    \end{subfigure}
    \caption{F1-Score, AUC and AVPR versus the number of anomalies in the test set for three different datasets.}
    \label{fig:accuracy_per_test_size}
\end{figure}

We analyse the effect of the contamination rate of the test set on the F1-score and AVPR metrics. To do so, we use a variant of Algorithm~\ref{algo:recycling} with a 20-80 train-test split on the clean samples only. We then re-inject a varying number of anomalous samples in the test set, from none to all of them. Figure~\ref{fig:accuracy_per_test_size} shows that F1-score and AVPR improve as more anomalies are added to the test set. Because the train set is fixed, this clearly shows that the F1-score and AVPR metrics are biased by the amount of anomalous samples in the test set. This sensitivity can be analysed theoretically.

\begin{figure}[t]
    \centering
    \includegraphics[width=\textwidth]{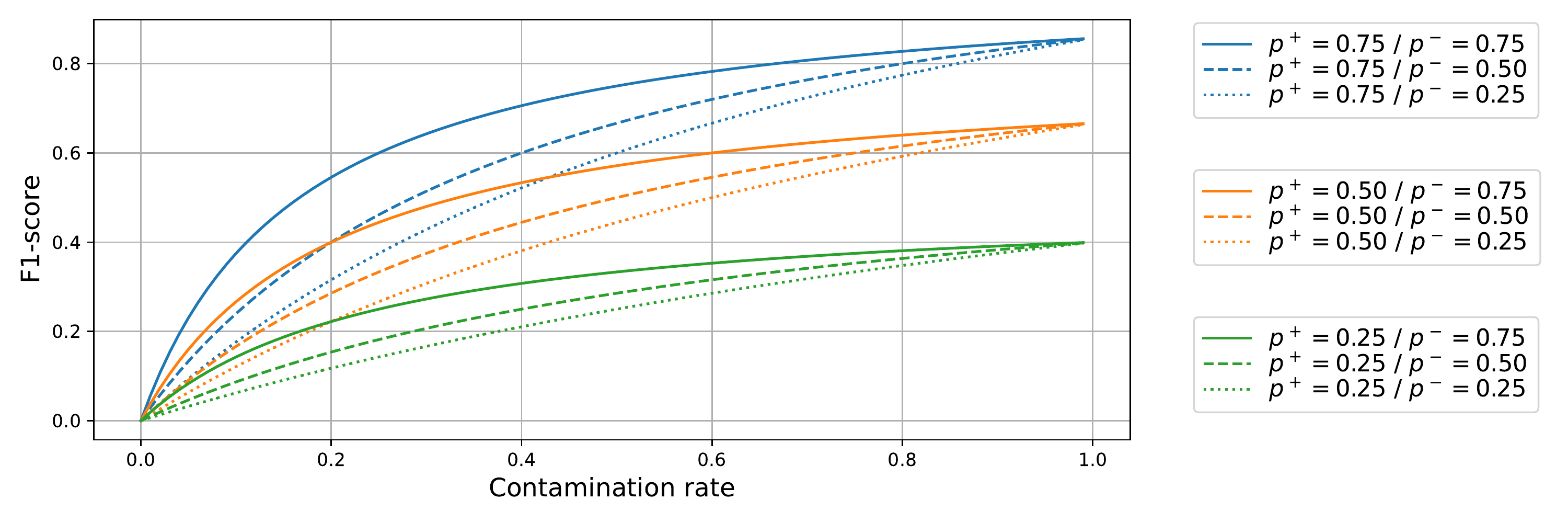}
    \caption{Theoretical F1-score for varying contamination rates of the test set, anomaly-detection capabilities $p^+$ and normal-detection capabilities $p^-$.}
    \label{fig:theory_F1}
\end{figure}

First, note that the contamination rate $\alpha = \frac{N_t^+}{N_t^++N_t^-} = \frac{N_t^+}{N_t^-}/\left(\frac{N_t^+}{N_t^-} + 1\right)$ is increasing with $\frac{N_t^+}{N_t^-}$. We start the analysis in a constant-threshold setting where the threshold $t$ does not depend on the test set, e.g. as in Algorithm~\ref{algo:theoretical}. In this setting, we can compute $p^- = \int_{\hat{s}=-\infty}^{t} P^-(\hat{s}) d\hat{s}$ the probability that the model classifies correctly a normal sample and $p^+ = \int_{\hat{s}=t}^{\infty} P^+(\hat{s}) d\hat{s}$ the probability that the model classifies correctly an anomalous sample (the recall). We observe that $tn = N_t^- * p^-$ and $fp = N_t^- * (1 - p^-)$ are directly proportional to $N_t^-$, while $tp = N_t^+ * p^+$ and $fn = N_t^+ * (1 - p^+)$ are directly proportional to $N_t^+$. As such, the recall $p^+$ does not depend on $\alpha$ while the precision (=$\frac{N_t^+ * p^+}{N_t^+ * p^+ + N_t^- * p^-} = \frac{\frac{N_t^+}{N_t^-} p^+}{\frac{N_t^+}{N_t^-} p^+ + p^-}$) increases with $\frac{N_t^+}{N_t^-}$ and therefore with $\alpha$. This proves the AVPR increases with $\alpha$ as the only value changing in Equation~\ref{eq:avpr} is the increasing precision. This also proves the F1-score with a fixed threshold is increasing with $\alpha$, as it is the harmonic mean of a constant and an increasing value. This theoretical variation of the F1-score is shown in Figure~\ref{fig:theory_F1}.

We now analyse the case where the threshold $t$ for the F1-score is computed using the test set as done in Algorithm~\ref{algo:recycling}. As we use a perfect estimation of the contamination rate, we have $recall=precision=\textit{F1-score}$. Let us analyse this quantity in the view of the recall and compare it to the constant-threshold setting. If we add an anomaly to the test set, there are two possibilities:
\begin{itemize}
    \item It is at the right side of the threshold, hence the threshold stays constant as there are still as many samples detected as anomalies as there are anomalies.
    \item It is at the wrong side of the threshold. The threshold therefore decreases to include one more sample as a predicted anomaly. There are two possibilities:
    \begin{itemize}
        \item This additional sample is an anomaly, in which case the recall increases, whereas it would have decreased in the constant-threshold setting.
        \item This additional sample is a clean sample, in which case the recall decreases the same way it would have decreased in the constant-threshold setting.
    \end{itemize}
\end{itemize}
Compared to the constant-threshold setting, the only difference is the case where the recall is better than expected thanks to the shift of the threshold. Therefore, adding anomalies increases the F1-score even more than in the constant-threshold setting, meaning the variable-threshold setting is even more biased by the contamination rate of the test set. More formally, if we add anomalies without changing the number of clean samples, the new threshold $t'$ will be smaller (or equal in the case of a perfect classifier) than the old one $t$, as we want to select more samples as being anomalies. The recall, precision and F1-score therefore increase from  $\int_{\hat{s}=t}^{\infty} P^+(\hat{s}) d\hat{s}$ (i.e. $p^+$ in the previous demonstration) to $\int_{\hat{s} = t'}^{t} P^+(\hat{s}) d\hat{s} + \int_{\hat{s}=t}^{\infty} P^+(\hat{s}) d\hat{s}$, which is greater or equal as a probability is always positive. Thus, if the classifier is not a perfect classifier, the F1-score increases with the contamination rate of the test set.

This concludes our demonstration that both the AVPR and the F1-score metrics are biased by the contamination rate of the test set.

\subsection{How to Artificially Increase your F1-Score and AVPR}

Combining the previous results and algorithms, we can define an algorithm to get an arbitrarily good F1-score or AVPR on any dataset. As shown in Section~\ref{subsec:sensitivity}, the F1-score and AVPR are sensitive to the contamination rate of the test set. Using the Algorithm~\ref{algo:recycling} from Section~\ref{subsec:train-test-split}, we can make this contamination rate vary. To do so, we only have to modify $\beta$, the amount of data used for the test set. Indeed, it modifies the number of normal samples $N_t^-$ in the test set while the number of anomalies $N_t^+$ stays the same. Pushed to the extreme, we can have near to no clean samples in the test set, resulting in a near-to-perfect F1-score and AVPR. This phenomenon is shown in Table~\ref{tab:cheat}. We can see that, by using the Algorithm~\ref{algo:recycling} the F1-score increases for all three datasets. This is because the anomalous sample of the train set are re-injected and thus the contamination rate of the test set increases. Then, using $5\%$ of the data for the test set instead of $20\%$ increase again the F1-score and AVPR.

\begin{table}[t]
    \centering
    \caption{Demonstration of the sensitivity of the metrics to the evaluation protocol. Optimal threshold is the threshold computed on the test set to obtain the best F1-score possible (unapplicable to AUC and AVPR).
    }
    \begin{tabular}{c c|c c c c}
        & Split procedure & Algo~\ref{algo:theoretical} & Algo~\ref{algo:recycling} & Algo~\ref{algo:recycling} & Algo~\ref{algo:recycling}\\
        & Test size & 20\% & 20\% & 5\% & 5\% \\
        & Threshold & estimated & estimated & estimated & optimal \\
        \hline
        \multirow{3}{*}{F1}&arrhythmia&\small 0.451\tiny($\pm$ 0.103)&\small 0.715\tiny($\pm$ 0.025)&\small 0.867\tiny($\pm$ 0.021)&\small 0.888\tiny($\pm$ 0.012)\\
        &kddcup&\small 0.102\tiny($\pm$ 0.025)&\small 0.762\tiny($\pm$ 0.004)&\small 0.940\tiny($\pm$ 0.002)&\small 0.971\tiny($\pm$ 0.001)\\
        &thyroid&\small 0.446\tiny($\pm$ 0.110)&\small 0.647\tiny($\pm$ 0.022)&\small 0.781\tiny($\pm$ 0.021)&\small 0.803\tiny($\pm$ 0.017)\\
        \hline
        \multirow{3}{*}{AVPR}&arrhythmia&\small 0.481\tiny($\pm$ 0.116)&\small 0.770\tiny($\pm$ 0.041)&\small 0.924\tiny($\pm$ 0.028)&\small 0.924\tiny($\pm$ 0.029)\\
        &kddcup&\small 0.299\tiny($\pm$ 0.017)&\small 0.653\tiny($\pm$ 0.015)&\small 0.872\tiny($\pm$ 0.008)&\small 0.873\tiny($\pm$ 0.007)\\
        &thyroid&\small 0.488\tiny($\pm$ 0.113)&\small 0.719\tiny($\pm$ 0.020)&\small 0.880\tiny($\pm$ 0.017)&\small 0.881\tiny($\pm$ 0.017)\\
        \hline
        \multirow{3}{*}{AUC}&arrhythmia&\small 0.809\tiny($\pm$ 0.065)&\small 0.806\tiny($\pm$ 0.020)&\small 0.803\tiny($\pm$ 0.042)&\small 0.799\tiny($\pm$ 0.042)\\
        &kddcup&\small 0.736\tiny($\pm$ 0.007)&\small 0.735\tiny($\pm$ 0.007)&\small 0.735\tiny($\pm$ 0.011)&\small 0.737\tiny($\pm$ 0.011)\\
        &thyroid&\small 0.935\tiny($\pm$ 0.027)&\small 0.931\tiny($\pm$ 0.005)&\small 0.929\tiny($\pm$ 0.009)&\small 0.929\tiny($\pm$ 0.009)\\
    \end{tabular}
    \label{tab:cheat}
\end{table}

\begin{figure}[t]
    \centering
    \includegraphics[width=\textwidth]{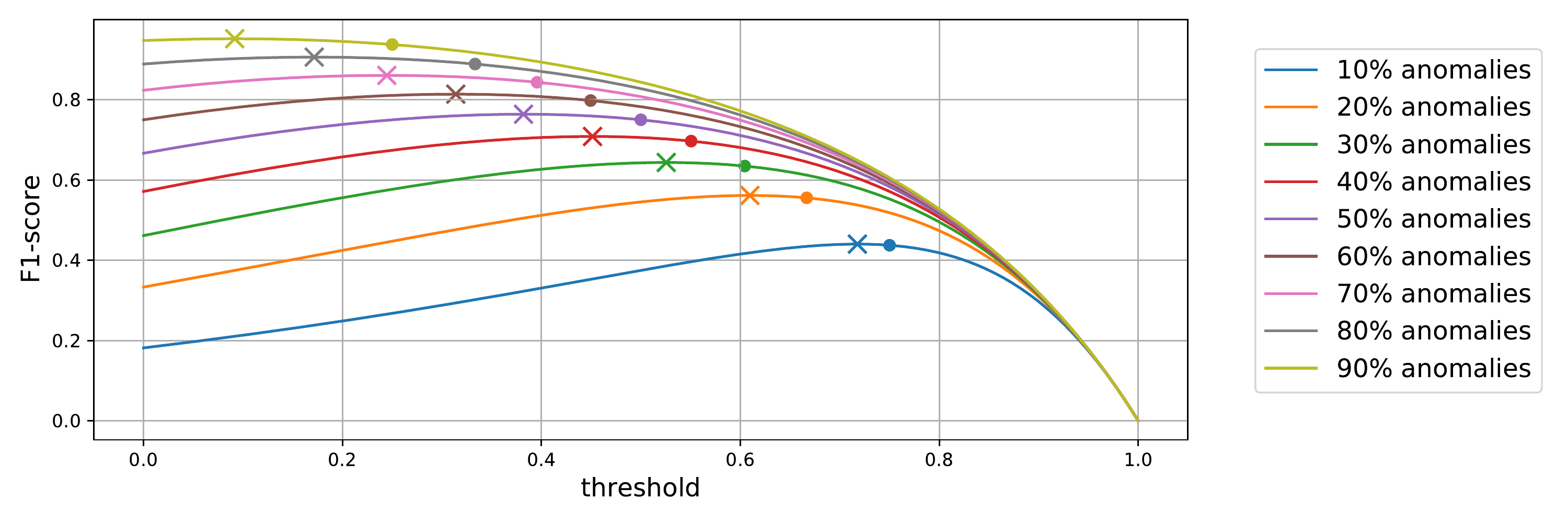}
    \caption{Theoretical example of the evolution of the F1-score for different thresholds and contamination rates of the test set. The model used is a toy model having $P^+(\hat{s}) = 2*\hat{s}$ and $P^-(\hat{s}) = 2 * (1 - \hat{s})$ for $0 \leq \hat{s} \leq 1$. Dots are the $fp = fn$ thresholds and crosses are the optimal thresholds.}
    \label{fig:theory_threshold}
\end{figure}

Another interesting observation is that fixing the threshold according to the contamination rate does not give the optimal F1-score~\cite{lipton2014}. In practice, using a threshold smaller than this one often results in a better F1-score, as visible in Figure~\ref{fig:precision_recall_f1} and shown in Figure~\ref{fig:theory_threshold}. As a consequence, we can artificially increase the F1-scores even more by computing the optimal threshold. This is shown in the last two columns of Table~\ref{tab:cheat}.

This proves that, with the exact same model and seemingly identical metrics, the F1-score can be greater and greater. This clearly supports the importance of specifying in detail the train-test split used and the way the threshold is computed. We observe in the literature that this part of the evaluation protocols is often missing or unclear \cite{9157105,zong2018deep,han2020gan,rvae}, and the reported results are therefore impossible to compare with. This is part of the reproducibility problem observed in the machine learning community. More importantly, some papers report results computed using different evaluation protocols \cite{mtq,han2020gan}, leading to meaningless comparisons that are nonetheless used to draw arbitrary conclusions.

\subsection{F1-Score Cannot Compare Datasets Difficulty}

\begin{figure}[t]
    \centering
    \begin{subfigure}{0.3\textwidth}
        \centering
        \includegraphics[width=\textwidth]{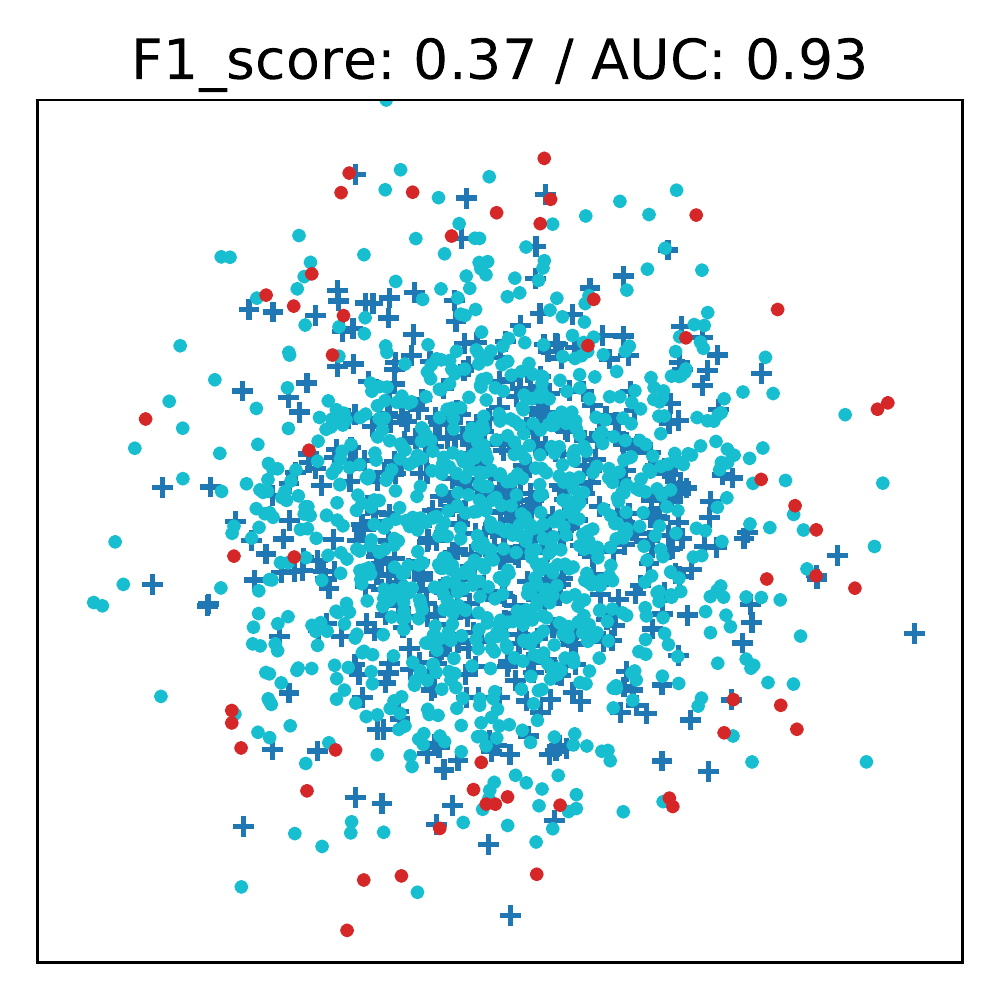}
        \caption{\centering Easy dataset \newline  $5\%$ contamination}
        \label{subfig:easy_dataset_5}
    \end{subfigure}%
    \begin{subfigure}{0.3\textwidth}
        \centering
        \includegraphics[width=\textwidth]{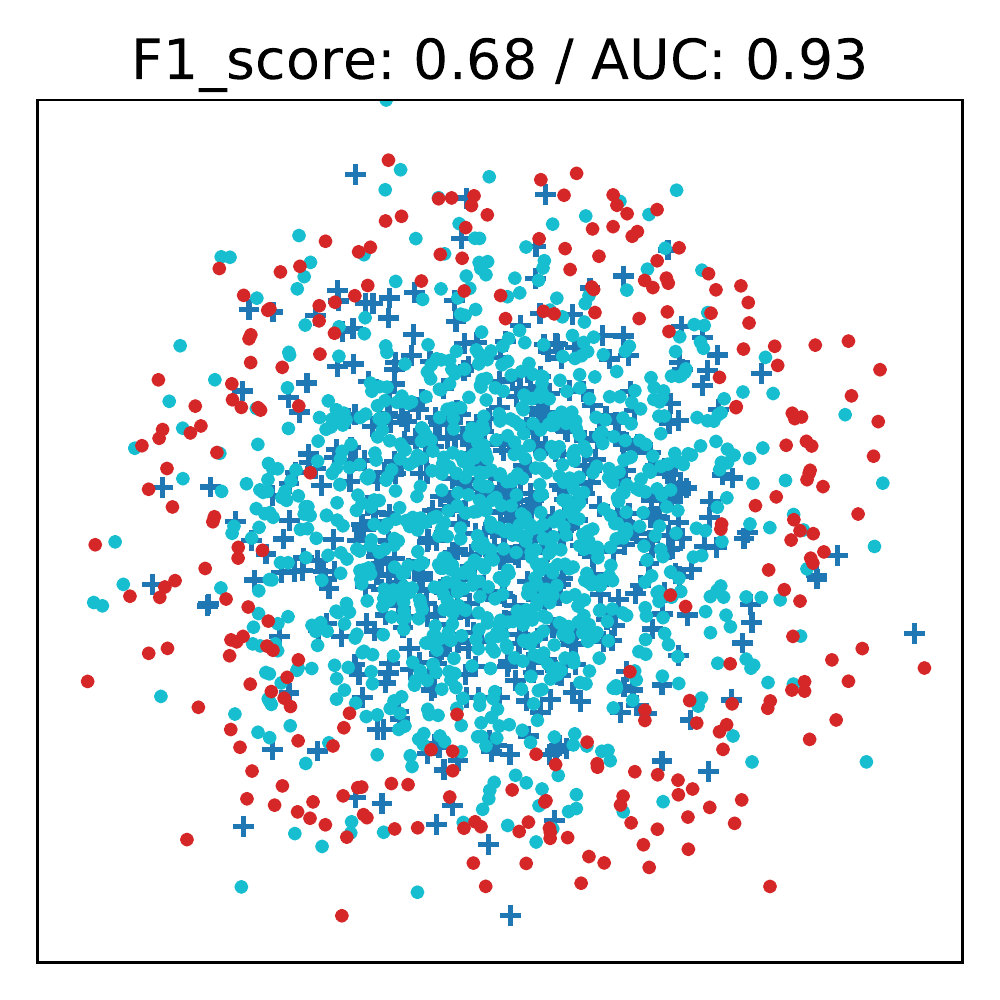}
        \caption{\centering Easy dataset \newline $20\%$ contamination}
        \label{subfig:easy_dataset_20}
    \end{subfigure}
    \begin{subfigure}{0.3\textwidth}
        \centering
        \includegraphics[width=\textwidth]{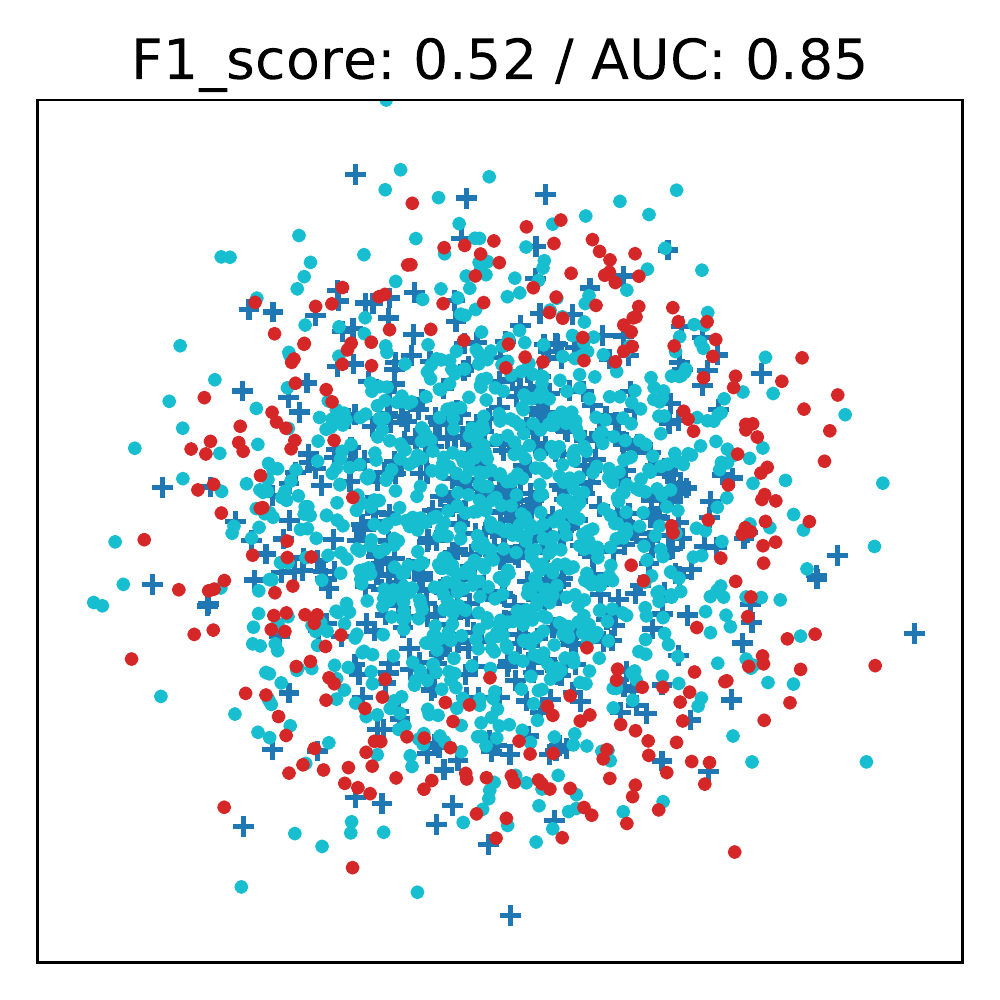}
        \caption{\centering Hard dataset \newline $20\%$ contamination}
        \label{subfig:hard_dataset_20}
    \end{subfigure}
    \begin{subfigure}{\textwidth}
        \centering
        \includegraphics[width=\textwidth]{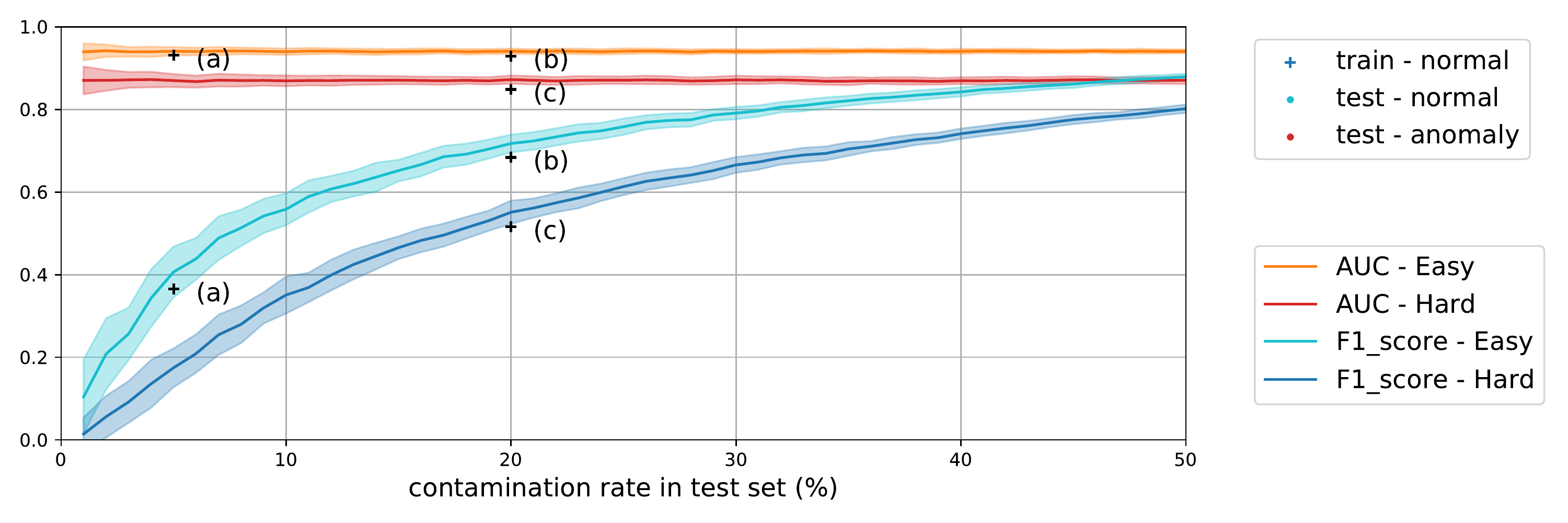}
        \caption{Performances comparison between easy toy dataset and hard toy dataset}
        \label{subfig:dataset_easy_hard}
    \end{subfigure}
    \caption{
    Analysis of the dataset comparison through different metrics. 
    We randomly draw normal samples from a standard gaussian distribution, and anomalous samples from a noisy a around the mean. By varying the radius of the circle - $2.5$ for the \textit{easy} case, $2.1$ for the \textit{hard} one - we change the difficulty of the dataset. The greater the radius, the easier it is to separate both distributions. A simple gaussian is used as model.}
    \label{fig:dataset_difficulty}
\end{figure}

Another shortcoming of the F1-score and AVPR metrics is the comparison between datasets. One may be tempted to conclude that a dataset on which an approach has a higher F1-score is easier to model than another dataset with a lower score. However, this intuition is flawed when using these metrics as they strongly depend on the contamination rate of these datasets.

Figure~\ref{fig:dataset_difficulty} highlights the dataset comparison problem.  
Figure~\ref{subfig:dataset_easy_hard} shows the F1-score and AUC obtained on two toy datasets, an easy one (with a big radius) and a hard one (with a small radius). We show that we can obtain a better F1-score on a hard dataset (Figure~\ref{subfig:hard_dataset_20}) than on an easy dataset (Figure~\ref{subfig:easy_dataset_5}) just by changing the contamination rate. With an equal contamination rate (Figure~\ref{subfig:easy_dataset_20}) we can see that the easy dataset is indeed easier to model.

This situation also appears in real-world datasets. Indeed, in Table~\ref{tab:cheat} with Algorithm~\ref{algo:theoretical}, the \textit{kdd cup} dataset appears harder than \textit{arrhythmia} and \textit{thyroid} as it obtains a worse F1-score. However, if we compare them with Algorithm~\ref{algo:recycling}, the \textit{kdd cup} dataset obtains better results than the other two. The comparison of the datasets difficulty is inconsistent and therefore unreliable.

\section{Call for Action}
\label{sec:proposed_alternative}
Given the instability shown in Section~\ref{sec:impact_setup_f1_score}, we suggest the anomaly-detection community to use the evaluation protocol described in Algorithm~\ref{algo:recycling} but using only the AUC metric. Other approaches could be adopted, but this one will give better comparability between reported results and these results will have lower variances.

\subsection{Use AUC}
We have demonstrated in Section~\ref{sec:impact_setup_f1_score} how the F1-score and AVPR metrics can be tricky to use and lead to wrong conclusions, slowing down the research in the field. To avoid these pitfalls, we recommend using the AUC metric. First of all, AUC is not sensitive to the contamination rate of the test set, as shown in Figure~\ref{fig:accuracy_per_test_size}. This can be proven by developing Equation~\ref{eq:auc}:
\begin{align}
    AUC &= \int_{t=-\infty}^{\infty} \frac{tp(t)}{tp(t) + fn(t)} \frac{d}{dt}\left(\frac{fp}{fp+tn}\right)\Bigr\rvert_{t} dt\\
    &= \int_{t=-\infty}^{\infty} \int_{\hat{s}=t}^{\infty} P^+(\hat{s}) d\hat{s} \frac{d}{dt}\left(\int_{\hat{s}'=-\infty}^{t} P^-(\hat{s}') d\hat{s}'\right)\Bigr\rvert_{t} dt\\
    % &= \int_{t=-\infty}^{\infty} \int_{\hat{s}=t}^{\infty} P^+(\hat{s}) P^-(t)~d\hat{s}~dt\\
    &= \int_{\{(\hat{s}, t) \in \mathbb{R}^2 \mid \hat{s} \geq t\}} P^+(\hat{s}) P^-(t)~d\hat{s}~dt \label{eq:AUC_symmetric}
\end{align}
which depends only on the model properties ($P^+$ and $P^-$) and not on the test set. This independence prevents most of the problems identified in the previous section. As illustrated in Figure~\ref{fig:dataset_difficulty}, datasets are more comparable using AUC. Moreover, Table~\ref{tab:cheat} highlights the stability of the AUC.

Additionally, there is no need to define a threshold when using AUC. This is a good thing as the choice of a threshold can prevent comparability. Indeed, most of the proposed models in the literature \cite{9157105,OCGAN,maziarka2020flowbased,zong2018deep,han2020gan} do not include a way to train a threshold. Therefore, arbitrary thresholds are used to compute the F1-score. The way to arbitrarily choose this threshold can vary from one paper to the other and lead to incomparable results. Even worse, this threshold could depend on the test set, such as the one producing $fp = fn$, thus having results biased by the contamination rate of the test set. This is not a problem with the AUC as it does not need a threshold. 

Finally, another source of non-comparability is the choice of the positive class. Some may choose the \textit{normal} class as positive \cite{DROCC,OCGAN} and other the \textit{anomaly} class as positive \cite{DBLP:journals/corr/abs-1904-00152,rpad,han2020gan}. AUC has the advantage of being independent of the choice of which class is seen as positive, as long as the scores are negated accordingly. Indeed, Equation~\ref{eq:AUC_symmetric} is symmetric between $P^+$ and $P^-$ up to the $\hat{s} \geq t$ part which is solved by negating the scores.

All in all, AUC is insensitive to many arbitrary choices in the evaluation protocol. It results in a better comparability between the different reported results.

\subsection{Do not Waste Anomalous Samples}

As, by definition, anomalous samples are rare, it is important to re-inject them in the test set, as described in Algorithm~\ref{algo:recycling}. Indeed, by using more anomalous samples in the test set, the variance in the metrics is lower. 

As shown in Table~\ref{tab:cheat}, when using AUC, Algorithm~\ref{algo:recycling} gives the same mean result than Algorithm~\ref{algo:theoretical}, but with a better precision (lower standard deviation). This is easily explained by the fact that there are more anomalies in the test set, increasing the applicability of the law of large numbers. This increased precision can be useful to obtain significant results rather than random-looking ones. Algorithm~\ref{algo:recycling} can be used as long as the metric used is not biased by the contamination rate of the test set. It is therefore compatible with the AUC metric.

\section{Conclusion}
The literature in the field of anomaly detection lacks precision in describing evaluation protocols. Because of the sensitivity of the F1-score and AVPR metrics to the contamination rate of the test set, this results in a reproducibility issue of the proposed works as well as a comparison problem between said works. Moreover, we observe that some works do the subtle mistake of comparing results produced with different evaluation protocols and draw arbitrary conclusions from it.
To solve this problem, we suggest the anomaly-detection community to use the AUC, which is insensitive to most arbitrary choices in the evaluation protocol. Moreover, we propose to use a \textit{recycling} algorithm (Algorithm~\ref{algo:recycling}) for the train-test split to make the most of anomalies in each dataset. These two actions will result in more comparable and more precise results across research teams.

% ---- Bibliography ----
% BibTeX users should specify bibliography style 'splncs04'.
% References will then be sorted and formatted in the correct style.
\bibliographystyle{splncs04}
\bibliography{bibliography}

\end{document}